\documentclass[sigconf]{acmart}[9pt]
\AtBeginDocument{%
  \providecommand\BibTeX{{%
    \normalfont B\kern-0.5em{\scshape i\kern-0.25em b}\kern-0.8em\TeX}}}

\copyrightyear{2024}
\acmYear{2024}
\setcopyright{acmlicensed}\acmConference[DAC '24]{61st ACM/IEEE Design Automation Conference}{June 23--27, 2024}{San Francisco, CA, USA}
\acmBooktitle{61st ACM/IEEE Design Automation Conference (DAC '24), June 23--27, 2024, San Francisco, CA, USA}
\acmDOI{10.1145/3649329.3657386}
\acmISBN{979-8-4007-0601-1/24/06}

\usepackage{url}
\usepackage{enumitem}

\usepackage[utf8]{inputenc} 
\usepackage[T1]{fontenc}    
\usepackage{url}            
\usepackage{booktabs}       
\usepackage{amsfonts}       
\usepackage{amsmath}
\usepackage{nicefrac}       
\usepackage{microtype}      
\usepackage{xcolor}         
\usepackage{colortbl}
\usepackage{graphicx}
\usepackage{svg}
\usepackage{multirow}
\usepackage{wrapfig}
\usepackage{threeparttable}
\usepackage{tikz}
\usepackage{makecell}
\usepackage{tablefootnote}

\usepackage{color}
\usepackage{xspace}
\usepackage{caption}
\usepackage{adjustbox}

\usepackage{subfigure}

\usepackage{amsmath}
\usepackage{mathtools}
\usepackage{amsthm}
\theoremstyle{plain}

\theoremstyle{definition}

\theoremstyle{remark}

\newcommand{\name}{HOGA\xspace}

\begin{document}

\title{Less is More: Hop-Wise Graph Attention for Scalable and Generalizable Learning on Circuits}


\author{Chenhui Deng$^1$, Zichao Yue$^1$, Cunxi Yu$^2$, Gokce Sarar$^3$, Ryan Carey$^3$, Rajeev Jain$^3$, Zhiru Zhang$^1$}
\affiliation{
  \institution{$^1$Cornell University, $^2$University of Maryland, $^3$Qualcomm Technologies, Inc.}
  \country{} 
}
\email{{cd574, zy383, zhiruz}@cornell.edu, cunxiyu@umd.edu, {gsarar, rcarey, rajeevj}@qti.qualcomm.com}

\renewcommand{\shortauthors}{Chenhui Deng, et al.}

\begin{abstract}
While graph neural networks (GNNs) have gained popularity for learning circuit representations in various electronic design automation (EDA) tasks, they face challenges in scalability when applied to large graphs and exhibit limited generalizability to new designs. These limitations make them less practical for addressing large-scale, complex circuit problems. In this work we propose HOGA, a novel attention-based model for learning circuit representations in a scalable and generalizable manner. HOGA first computes hop-wise features per node prior to model training. Subsequently, the hop-wise features are solely used to produce node representations through a gated self-attention module, which adaptively learns important features among different hops without involving the graph topology. As a result, HOGA is adaptive to various structures across different circuits and can be efficiently trained in a distributed manner. To demonstrate the efficacy of HOGA, we consider two representative EDA tasks: quality of results (QoR) prediction and functional reasoning. Our experimental results indicate that (1) HOGA reduces estimation error over conventional GNNs by $46.76\%$ for predicting QoR after logic synthesis; (2) HOGA improves $10.0\%$ reasoning accuracy over GNNs for identifying functional blocks on unseen gate-level netlists after complex technology mapping; (3) The training time for HOGA almost linearly decreases with an increase in computing resources.
Source code of HOGA is freely available at: \href{https://github.com/cornell-zhang/HOGA}{github.com/cornell-zhang/HOGA}.
\end{abstract}

\maketitle

\section{Introduction}

Recent years have seen a surge of interest in machine learning (ML) for electronic design automation (EDA),
which holds great potential in achieving faster design closure and minimizing the need for extensive human supervision~\cite{huang-mleda-survey-todaes2021}.
In particular, graph neural networks (GNNs) have become increasingly popular in the EDA community due to their ability to 
encode graph-structured data such as gate-level netlists into compact representations, which can be used for a multitude of downstream EDA applications, including quality of results (QoR) prediction and functional reasoning~\cite{sanchez2023comprehensive, wu2023gamora}.

\begin{figure}[t!]
\begin{center}
\centerline{\includegraphics[width=0.9\columnwidth]{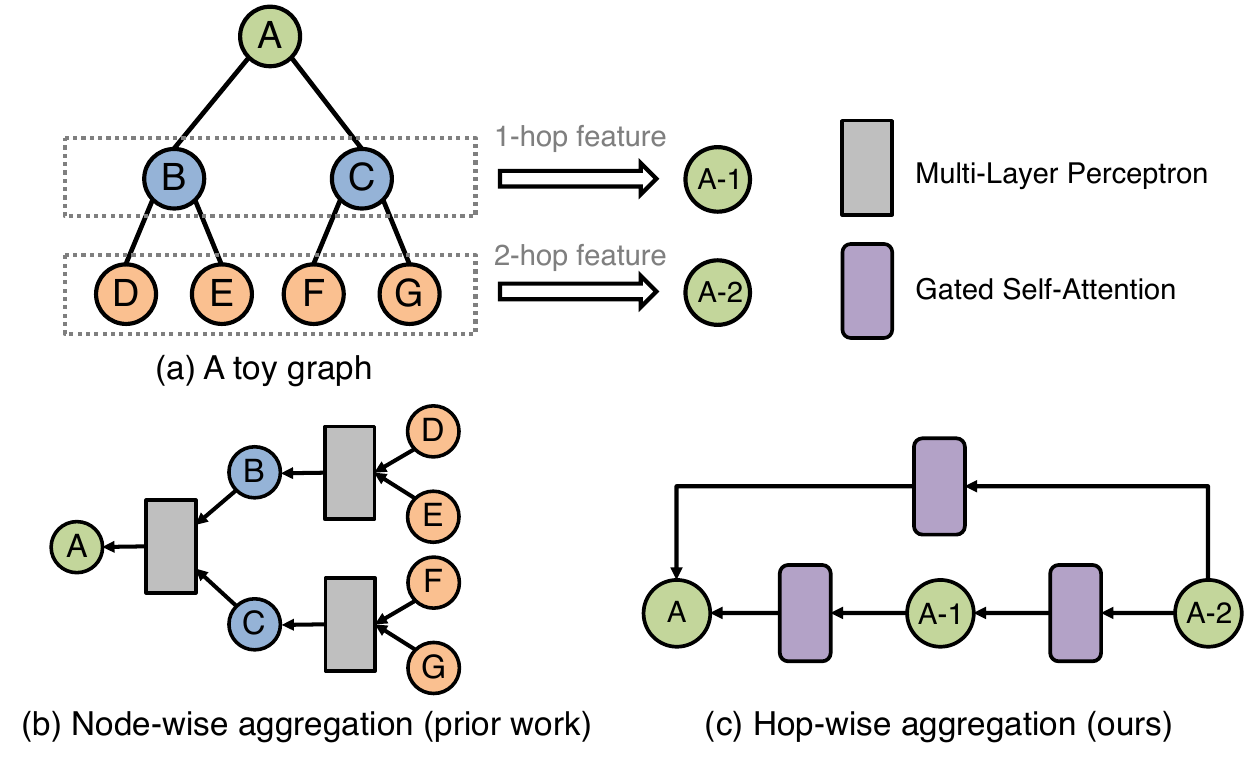}}
\vspace{-10pt}
\caption{Comparison of \name and prior GNNs --- (a) An example graph for illustration; (b) GNN computation graph; (c) Computation graph of our proposed approach, \name.}
\label{figure:toy_example}
\end{center}
\vspace{-30pt}
\end{figure}

However, scaling GNN training to large graphs is a notoriously challenging problem, which poses a serious concern on the practical benefit of GNNs on large-scale EDA problems. On the one hand, unlike common datasets on social networks and molecular graphs, which consist of either a few large graphs or a large number of small graphs, the circuit datasets may contain numerous large graphs. For instance, the OpenABC-D benchmark provides $870$k gate-level netlists, where each netlist consists of up to $240$k logic gates~\cite{chowdhury2021openabc}. Thus, training GNNs on such a large-scale circuit dataset is even more challenging than other graph-based applications. 
On the other hand, modern GNN models are built upon a message-passing paradigm, which learns representations through a recursive node-wise aggregation scheme 
shown in Figure \ref{figure:toy_example}(b).
As a consequence, it is nontrivial to perform efficient distributed GNN training due to the node dependencies in a graph structure.

Apart from the scalability challenge, it is also underexplored how to make GNNs generalizable across different circuit designs. Although there are many customized GNNs previously proposed for various EDA applications, their model backbones mainly follow classic GNNs such as GCN~\cite{kipf2016semi} and GraphSAGE~\cite{hamilton2017inductive}, which are not necessarily suitable for circuit problems. 
Consider a task of identifying functional blocks within circuits~\cite{wu2023gamora}. As distinct functional blocks may have different depths, the number of hops to be considered varies across nodes, which cannot be easily captured by common GNNs. 
Moreover, the high-order structures of functional blocks are also important yet ignored by the aforementioned GNN models. As a result, existing GNNs for EDA tasks often  struggle to learn the intrinsic and critical information from complex circuit graphs, resulting in limited generalizability to unseen designs.

In literature, improving the scalability and enhancing the generalizability of GNNs on circuits are largely viewed as two orthogonal directions and seldomly explored together. 
Previous GNNs either incur massive communication overhead among GPUs, or rely on heuristic graph sampling algorithms that may lose critical structural information~\cite{shao2022distributed}, which in turn degrades the model generalizability. In the context of learning generalizable GNNs for circuit problems, Wang et al.~\cite{wang2022functionality} leveraged graph contrastive learning to pretrain GNNs and adopted multiple neighbor aggregation functions to learn the functionalities of logic gates, which comes with additional computational costs and thus worsens the model scalability.


Motivated by the limitations of conventional GNNs on circuit designs, we propose a novel hop-wise attention approach, named \name, to improve both the scalability and generalizability of circuit representation learning.
As shown in Figure \ref{figure:toy_example}(c), \name adopts a hop-wise aggregation scheme, which precomputes the hop-wise features and only uses those features to learn node representations through a gated self-attention module.
Since the final representation per node depends solely on its own hop-wise features, there are no dependencies between different nodes during training, 
making it easy to scale \name training in a distributed manner.
Moreover, 
the gated self-attention module enables \name to adaptively capture critical high-order structures from different hops per node, rendering it generalizable across different circuit designs. 

Notably, \name is a flexible approach for learning circuit representations, which can be integrated with other customization techniques previously proposed for various downstream EDA tasks.
To demonstrate the viability and flexibility of \name, we consider two representative circuit problems: (1) QoR prediction -- We focus on predicting the optimized gate count after logic synthesis on the OpenABC-D benchmark, which is generated from various circuit designs as well as synthesis recipes, and is one of the largest open-sourced circuit datasets; (2) Functional reasoning -- We follow the most challenging setting in Gamora~\cite{wu2023gamora} by identifying functional blocks on gate-level netlists after technology mapping. Our experiments show that \name not only outperforms prior GNNs on unseen designs, but is also efficient for distributed training. We summarize our main technical contributions as follows:

\noindent $\bullet$ To the best of our knowledge, we are the first to introduce a scalable and generalizable model for circuit representation learning, which is achieved by a novel hop-wise graph attention scheme.

\noindent $\bullet$ By precomputing hop-wise features, \name avoids recursive neighbor aggregation, which enables \name to learn each node representation independently and facilitates massive parallelization for distributed training. As a result, the training time of \name almost linearly decreases with an increase in computing resources.


\noindent $\bullet$ Owing to the proposed gated self-attention module, \name is able to adaptively learn critical and high-order circuit structures, leading to $46.76\%$ error drop and $10.0\%$ accuracy improvement for QoR prediction and functional reasoning on new designs, respectively.

\section{Preliminary and Motivation}
\subsection{Graph Neural Networks at Scale}
\label{scale_motivation}
GNNs have emerged as a promising technique that encodes circuit graphs into compact representations, which can then be utilized for addressing a wide spectrum of EDA problems. Specifically, given a circuit graph $\mathcal{G}=(\mathcal{V}, \mathcal{E})$,
GNNs update initial node features $\{x_i^{(0)} \; | \; i \in \mathcal{V}\}$ based on a node-wise aggregation scheme as follows:
\begin{equation}\label{eqn:agg}
    m_i^{(l)} = f^{(l)}(\{x_j^{(l-1)} \; | \; j \in \mathcal{N}(i)\}), \; x_i^{(l)} = g^{(l)}(m_i^{(l)}, x_i^{(l-1)})
    \vspace{-5pt}
\end{equation}
where $x_i^{(l)}$ denotes the node feature vector at the $l$-th layer, $\mathcal{N}(i)$ represents the set of neighbors of node $i$, $f$ can be any permutation-invariant function (e.g., mean-pooling), and the goal of function $g$ is to update node representations based on the aggregated features from neighbors. After stacking $L$ GNN layers, the output features $\{x_i^{(L)} \; | \; i \in \mathcal{V}\}$ serve as the final node representations.
However, let $d$ be the average node degree, there are $O(d^L)$ nodes required to obtain $x_i^{(L)}$, which increases exponentially with the number of layers and is known as the ``neighbor explosion'' issue. 
Besides, 
the graph dependencies also incur significant communication overhead and work imbalance, hindering efficient distributed GNN training. To tackle both challenges, 
numerous efforts have been devoted to improve GNN training efficiency by adopting different sampling strategies~\cite{hamilton2017inductive, chen2018fastgcn, zeng2019graphsaint}. 
While these methods demonstrate promising results
on social networks, we argue that they are unsuitable for circuits as the sampling algorithms may entirely break the design functionality and lead to poor accuracy. This is empirically confirmed in Section \ref{eval_reason}.
In contrast, we introduce a hop-wise aggregation scheme that independently learns each node representation based on its own hop-wise features, which renders our model embarrassingly parallel and greatly facilitates distributed training.

\vspace{-8pt}
\subsection{Generalizable Graph Learning in EDA}
\label{gen_gnn}
Since realistic circuit designs may originate from distinct domains, generalization capability is crucial for deploying graph learning models on circuits. 
To this end, Ustun et al. proposed a customized GNN model by distinguishing predecessors and successors in a graph, which demonstrates promising generalizability on learning operation mapping patterns~\cite{ustun2020dsage}. Later, Zhang et al.~\cite{zhang2020grannite} and Guo et al.~\cite{guo2022timing} introduced customized GNNs tailored for power inference and timing prediction tasks respectively, via sequentially updating node representations. More recently, 
Wu et al. presented a multi-task graph learning framework for functional reasoning. 
While these methods showcase promising results on the respective tasks, their GNN backbones are built upon the conventional message-passing paradigm, which cannot capture critical high-order structures formed by multiple nodes. Notably, many functional blocks (e.g., full adders) are essentially high-order structures, which are crucial for many circuit problems such as functional reasoning~\cite{wu2023gamora}. 
Therefore, the aforementioned methods still cannot capture the intrinsic circuit information, leading to either restrictions to specific tasks or limited generalizability to complex circuit designs.

Although Wang et al. attempted to improve GNN generalizability by 
adopting the notion of graph contrastive learning~\cite{wang2022functionality}, it involves nontrivial computational costs and thus worsens the GNN scalability. In this work, we aim to devise a graph learning model that is both scalable and generalizable for circuit designs by using a gated self-attention module on hop-wise features per node.

\vspace{-5pt}
\section{The Proposed Approach}
\begin{figure*}[t!]
\begin{center}
	\includegraphics[width=0.9\textwidth]{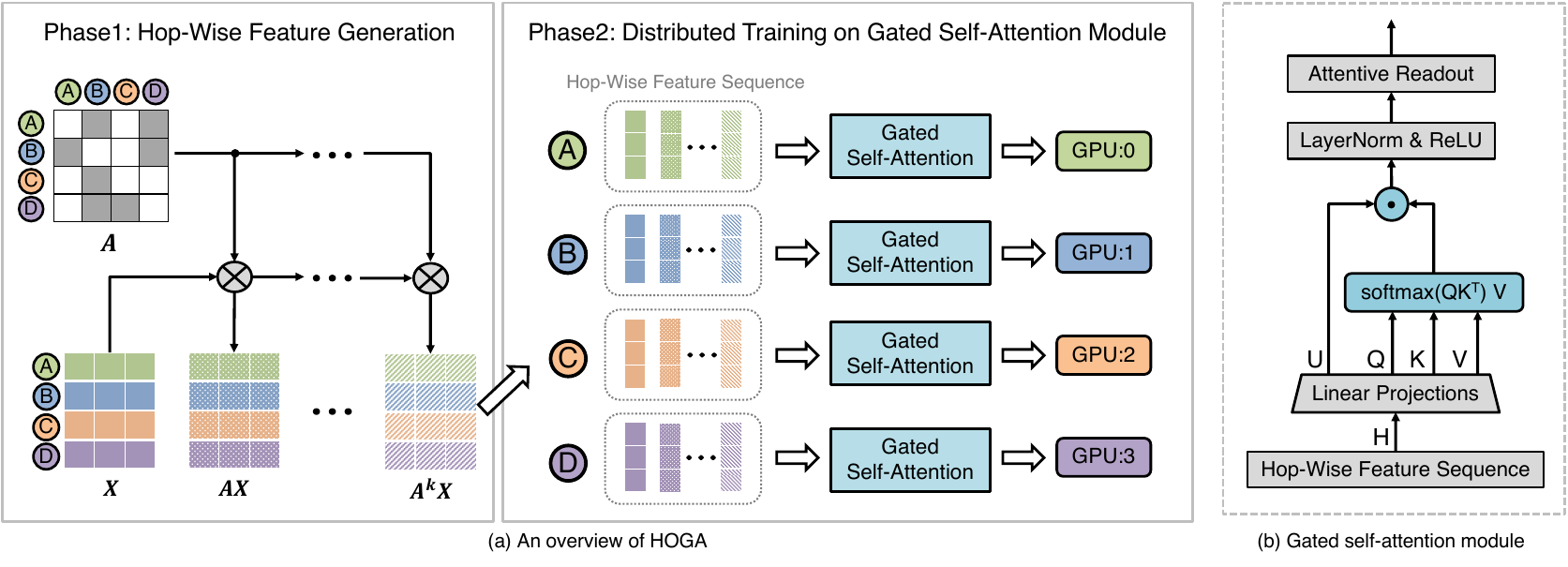}
 \vspace{-15pt}
	\caption{(a) An overview of the two major phases of \name; (b) Our introduced gated self-attention module.} 
 \protect\label{figure:overview}
\end{center}
\vspace{-15pt}
\end{figure*}
\textbf{Problem formulation.} 
Given a graph adjacency matrix $A$ and a node feature matrix $X$,
our goal is to build a model $\mathcal{M}$ for learning high-quality node representations $Y = \mathcal{M}(A, X)$, which can be utilized for a broad spectrum of circuit problems.

Figure \ref{figure:overview}(a) gives an overview of \name that consists of two major phases. During the first phase, \name computes hop-wise features by iteratively multiplying the adjacency matrix with the node feature matrix, as described in Section \ref{preprocess}. 
Note that this step can be finished in advance, 
allowing \name training with complexity independent of the graph structure.
The second phase trains a gated self-attention module to capture high-order interactions among hop-wise features, as illustrated in Section \ref{gated_attn}. 
Consequently, \name is generalizable to different circuit designs while simultaneously benefiting from high parallelism for distributed training.

\vspace{-5pt}
\subsection{Hop-Wise Feature Generation}
\label{preprocess}
As depicted in Figure \ref{figure:toy_example}(b), traditional message-passing GNNs recursively aggregate features from neighbors, which leads to their poor scalability on large graphs. In contrast, we adopt a coarse-grained message-passing scheme based on hop-wise feature aggregation. To this ends, our first step is to generate hop-wise features.

Given the adjacency matrix $A \in \mathbb{R}^{n\times n}$ and node feature matrix $X \in \mathbb{R}^{n\times d}$, where $n$ and $d$ represent the number of nodes and the feature dimension respectively, we first normalize the adjacency matrix: $\hat{A} = D^{-\frac{1}{2}}AD^{-\frac{1}{2}}$, where $D$ is the node degree matrix. Next, we generate hop-wise features by iteratively computing Equation \eqref{eqn:hop_feats}, where 
$X^{(0)}=X$ and $K$ denotes the number of hops. 
\begin{equation}\label{eqn:hop_feats}
    X^{(k)} = \hat{A}X^{(k-1)}, \; k=1, 2, ..., K
\end{equation}

After obtaining hop-wise features $X^{(0)}, X^{(1)}, ..., X^{(K)}$, we stack them to construct a third-order tensor $\mathcal{X} \in \mathbb{R}^{n \times (K+1) \times d}$ such that:
\begin{equation}\label{eqn:hop_seq}
    \mathcal{X}_i = [X_i^{(0)}, X_i^{(1)}, ..., X_i^{(K)}]^T, \; i=1, 2, ..., n
\end{equation}
Consequently, for each $i \in \{1, 2, ..., n\}$, $\mathcal{X}_i$ comprises up to $K$-hop features of node $i$, which are then independently used to learn the corresponding node representation $Y_i$. Hence, there are no dependencies between different nodes,
making it easy to scale \name through distributed training. 
It is noteworthy that the time required for generating hop-wise features is generally negligible compared to the overall training time, as empirically confirmed in Section \ref{eval_qor}. 

While there are a few prior arts (e.g. SIGN~\cite{frasca2020sign}) augmenting node features by adopting a similar approach to Equation \eqref{eqn:hop_feats}, they simply train a multi-layer perceptron (MLP) model on augmented features. In contrast, we consider learning node representations through a hop-wise feature aggregation scheme, which is built upon a novel gated self-attention module introduced in the following section. 


\subsection{Hop-Wise Gated Attention}
\label{gated_attn}
Since \name takes $\mathcal{X}_i \in \mathbb{R}^{(K+1) \times d}$ as input and produces a representation $Y_i \in \mathbb{R}^{d}$ for every node $i$ independently, we omit the node index $(i)$ in the ensuing discussion, by simply denoting $\mathcal{X}_i$ as $H$ and $Y_i$ as $y$ for clarity.
A straightforward way of producing $y$ is to accumulate hop-wise features in $H$, i.e., $y = \sum_{k=0}^K H_k$.
However, this approach has two 
flaws: (1) it fails to capture high-order feature interactions among different hop neighbors, resulting in its limited expressivity of learning high-order circuit structures.
(2) it uniformly combines features from different hops and thus cannot identify and focus on important hop-wise features per node. 
To address those limitations, 
let us first consider a simple gated layer:
\vspace{-3pt}
\begin{equation}\label{eqn:gate}
U = HW_U, V=HW_V, \; \hat{H} = U \odot V
\vspace{-3pt}
\end{equation}
where $W_U, W_V \in \mathbb{R}^{d \times d}$ are trainable weight matrices, and $\odot$ denotes the element-wise product. We can derive from Equation \eqref{eqn:gate} that $\hat{H}_k$ captures the second-order interaction $(H_kW_U) \odot (H_kW_V)$ for every hop $k \in \{0, 1, ..., K\}$. 
However, this also means Equation \eqref{eqn:gate} fails to capture interactions among different hop-wise features, i.e., $(H_kW_U) \odot (H_jW_V)$ with $k \neq j$.
To tackle this issue, we introduce a gated self-attention layer in the following:
\vspace{-3pt}
\begin{equation}\label{eqn:gated_self_attn}
S = softmax(QK^T), \; \hat{H} = U \odot (SV)
\vspace{-3pt}
\end{equation}
where 
$Q=HW_Q, K=HW_K$, $W_Q, W_K \in \mathbb{R}^{d \times d}$ are trainable weight matrices, and $S$ is the self-attention matrix widely used in Transformer~\cite{vaswani2017attention}. Based on Equations \eqref{eqn:gate} and \eqref{eqn:gated_self_attn}, we can derive that $\hat{H}_k = (H_kW_U) \odot (\sum_{j=0}^K S_{k,j}H_jW_V) = \sum_{j=0}^K S_{k,j} \; (H_kW_U) \odot (H_jW_V)$, which captures second-order interactions on different hop-wise features. By stacking more layers, the output $\hat{H}$ naturally captures higher-order feature interactions from different hops, which correspond to higher-order structures on the input circuit graph.


\begin{figure*}[t!]
\begin{center}
	\includegraphics[width=0.9\textwidth]{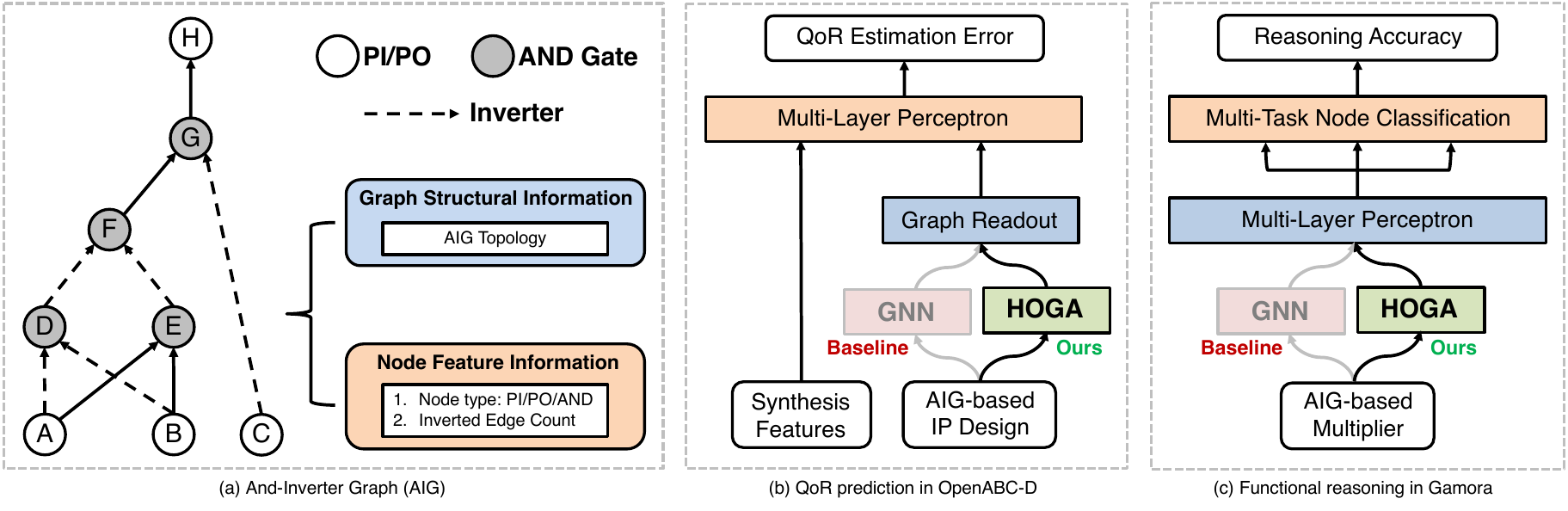}
        \vspace{-13pt}
	\caption{(a) An example of And-Inverter Graph, where each node represents a primary input/output (PI/PO) or an AND gate. The dashed arrows denote inverted edges; (b) An overview of QoR prediction in OpenABC-D; (c) An overview of functional reasoning in Gamora --- We solely replace GNNs with HOGA for learning circuit representations in both tasks. 
 } 
 \protect\label{figure:setup}
\end{center}
\vspace{-12pt}
\end{figure*}
\textbf{Implementation details.} To improve the training stability of \name, we add LayerNorm and ReLU to Equation \eqref{eqn:gated_self_attn} in our implementation, i.e., $\hat{H} = \text{ReLU}(\text{LayerNorm}(U \odot (SV)))$.
After obtaining $\hat{H} = [\hat{H}_0, \hat{H}_1, ..., \hat{H}_K]^T \in \mathbb{R}^{(K+1) \times d}$, we adopt an attentive readout scheme
to produce the final node representation $y$:
\begin{equation}\label{eqn:adapt_attn}
\begin{split}
c_k = \frac{exp(\alpha^T(\hat{H}_0||\hat{H}_k))}{\sum_{j=1}^K exp(\alpha^T(\hat{H}_0||\hat{H}_j))}, \; y = \hat{H}_0 + \sum_{k=1}^K c_k \hat{H}_k
\end{split}
\end{equation}
where $\alpha \in \mathbb{R}^{2d}$ is a trainable vector, $||$ denotes the concatenation operator, and $c_k$ represents an attention score to measure the importance of the $k$-hop feature $\hat{H}_k$ to the final node representation $y$. In this way, \name can identify and adaptively aggregate critical features from different hops to produce high-quality node representations, which are then used for downstream circuit tasks.

\subsection{Complexity analysis of \name}
Suppose we consider $h$ feature dimensions and $K$ hops. The complexity of hop-wise feature generation is $O(Kh|\mathcal{E}|)$. Besides, the gated self-attention module has a complexity of $O(Kh^2|\mathcal{V}| + K^2h|\mathcal{V}|)$ for linear projections and computing Equation \eqref{eqn:gated_self_attn}. Therefore, the total complexity of \name is $O(Kh|\mathcal{E}| + Kh^2|\mathcal{V}| + K^2h|\mathcal{V}|) = O(|\mathcal{E}| + |\mathcal{V}|)$, which is linear with respect to the number of nodes/edges.



\vspace{-5pt}
\section{Experiment}
\subsection{Experimental Setup}
We evaluate \name on two tasks: (1) \textbf{QoR prediction.} We focus on the OpenABC-D benchmark for predicting the optimized gate count in And-Inverter-Graphs (AIGs) after logic synthesis optimization through ABC~\cite{brayton2010abc}. Notably, OpenABC-D consists of $870,000$ AIGs that are generated by running various synthesis recipes on IPs 
from MIT LL labs CEP~\cite{brendon2019cep}, OpenCores~\cite{opencore}, and IWLS~\cite{albrecht2005iwls}. 
To demonstrate the generalizability of our approach, we train \name on the top $20$ designs in Table \ref{statistics} and evaluate it on the rest of the designs;
(2) \textbf{Functional reasoning.} We follow Gamora~\cite{wu2023gamora} to identify adder blocks by predicting the \textit{sum} and \textit{carry-out} nodes in AIGs of carry-save array (CSA) and Radix-$4$ Booth multipliers.
As technology mapping can largely increase functional reasoning complexity~\cite{li2013wordrev, yu2016formal}, we consider the most challenging scenario in Gamora, 
where AIGs are generated by ABC with complex ASAP $7$nm technology mapping, and
we only train \name on an $8$-bit multiplier design and perform inference on multipliers with bitwidth up to $768$.

Note that the authors of OpenABC-D and Gamora have proposed their own GNN-based models for the aforementioned tasks.
To ensure a fair comparison, we only replace their GNN blocks with \name and keep other model components the same as shown in Figures \ref{figure:setup}(b) and \ref{figure:setup}(c). 
In regard to HOGA hyperparameter settings, we adopt Adam optimizer with a learning rate of $0.0001$, a hidden dimension of $256$, and fix the number of gated self-attention layer to $1$. Besides, we set the number of hops $K$ as $5$ for experiments on OpenABC-D and $8$ on Gamora, which captures the information from the same number of hops as the baseline GNNs in both tasks. 
We leverage \textit{DistributedDataParallel} in PyTorch for distributed training on \name. All experiments are conducted on a Linux machine with an Intel Xeon Gold $5218$ CPU and 4 RTX A$6000$ GPUs.

\begin{table}[t!]
\centering
\caption{Statistics of OpenABC-D benchmark --- The top and bottom designs are used for training and test, respectively.}
\vspace{-5pt}
\label{statistics}
\scalebox{0.85}{
\begin{tabular}[t]{lrrl}\toprule
\textbf{IP Design}            &   \textbf{Nodes}   & \textbf{Edges}   & \textbf{Category}    \\ 
\midrule
spi &  $4219$ &  $8676$   &  Communication \\
i2c &  $1169$ &  $2466$   &  Communication \\
ss\_pcm &  $462$ &  $896$   &  Communication \\
usb\_phy &  $487$ &  $1064$   &  Communication \\
sasc &  $613$ &  $1351$   &  Communication \\
wb\_dma &  $4587$ &  $9876$   &  Communication \\
simple\_spi &  $930$ &  $1992$   &  Communication \\
pci &  $19547$ &  $42251$   &  Communication \\
dynamic\_node &  $18094$ &  $38763$   &  Control \\
ac97\_ctrl &  $11464$ &  $25065$   &  Control \\
mem\_ctrl &  $16307$ &  $37146$   &  Control \\
des3\_area &  $4971$ &  $10006$   &  Crypto \\
aes &  $28925$ &  $58379$   &  Crypto \\
sha256 &  $15816$ &  $32674$   &  Crypto \\
fir &  $4558$ &  $9467$   &  DSP \\
iir &  $6978$ &  $14397$   &  DSP \\
idft &  $241552$ &  $520523$   &  DSP \\
dft &  $245046$ &  $527509$   &  DSP \\
tv80 &  $11328$ &  $23017$   &  Processor \\
fpu &  $29623$ &  $59655$   &  Processor \\
\midrule
wb\_conmax &  $47840$ &  $97755$   &  Communication \\
ethernet &  $67164$ &  $144750$   &  Communication \\
bp\_be &  $82514$ &  $173441$   &  Control \\
vga\_lcd &  $105334$ &  $227731$   &  Control \\
aes\_xcrypt &  $45840$ &  $93485$   &  Crypto \\
aes\_secworks &  $40778$ &  $84160$   &  Crypto \\
jpeg &  $114771$ &  $234331$   &  DSP \\
tiny\_rocket &  $52315$ &  $108811$   &  Processor \\
picosoc &  $82945$ &  $176687$   &  Processor \\
\bottomrule
\end{tabular}
}
\vspace{-10pt}
\end{table}
\begin{table*}[ht!]
\centering
\caption{Comparison of \name and GCN for QoR prediction --- We choose mean absolute percentage error (MAPE) as the evaluation metric (Lower score is better). \name-$2$ and \name-$5$ indicate $K=2$ and $K=5$ in \name, respectively.}
\label{qor}
\vspace{-10pt}
\renewcommand{\arraystretch}{1} 
\begin{adjustbox}{width=\textwidth,center}
\begin{tabular}[t]{lrrrrrrrrr|rl}\toprule
           &   wb\_conmax  &  ethernet &  bp\_be   & vga\_lcd   & aes\_xcrypt   & aes\_secworks  & jpeg & tiny\_rocket  & picosoc & Average  & Training Time    \\ 
\midrule
GCN  &   $24.66\%$   &   $19.20\%$  &  $39.53\%$   &   $52.35\%$  &    $21.41\%$   &   $23.93\%$  &  $22.11\%$   &   $19.89\%$  &   $10.51\%$ &   $26.0\%$ &   $11.9$ hours ($1.0 \times$)  \\
\name-$2$  &   $7.33\%$   &   $6.51\%$  &  $17.81\%$   &   $9.31\%$  &    $\mathbf{5.79\%}$   &   $10.17\%$  &  $\mathbf{3.87\%}$   &   $6.70\%$  &   $2.91\%$ &   $7.8\%$ &   $\mathbf{3.8}$ hours ($\downarrow 3.1 \times$)  \\
\name-$5$  &   $\mathbf{4.53\%}$   &   $\mathbf{4.21\%}$  &  $\mathbf{4.76\%}$   &   $\mathbf{5.59\%}$  &    $6.57\%$   &   $\mathbf{8.42\%}$  &  $5.65\%$   &   $\mathbf{3.88\%}$  &   $\mathbf{1.90\%}$ &  $\mathbf{5.0\%}$ &  $11.2$ hours ($\downarrow 1.1 \times$)  \\
\bottomrule
\end{tabular}
\end{adjustbox}
\end{table*}
\begin{figure*}[ht]
    \centering
    \vspace{-10pt}
    \subfigure[]{%
        \includegraphics[width=0.24\textwidth]{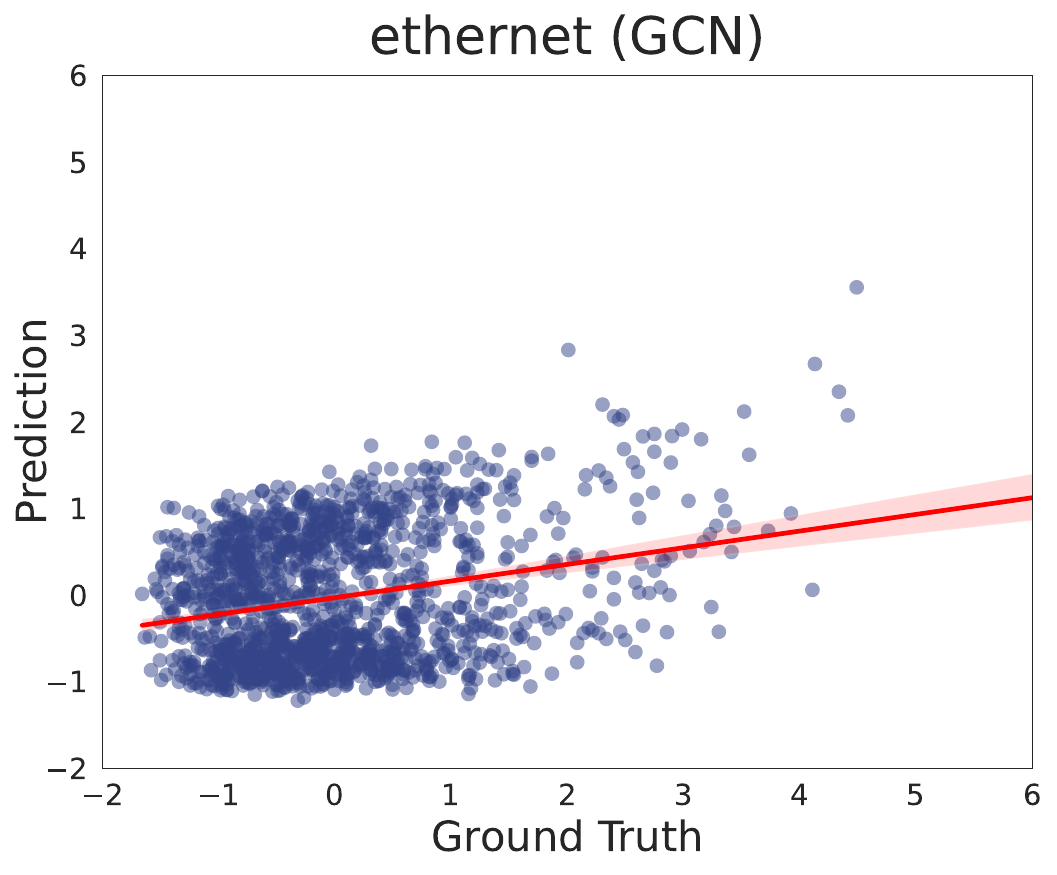}}
    \subfigure[]{%
        \includegraphics[width=0.24\textwidth]{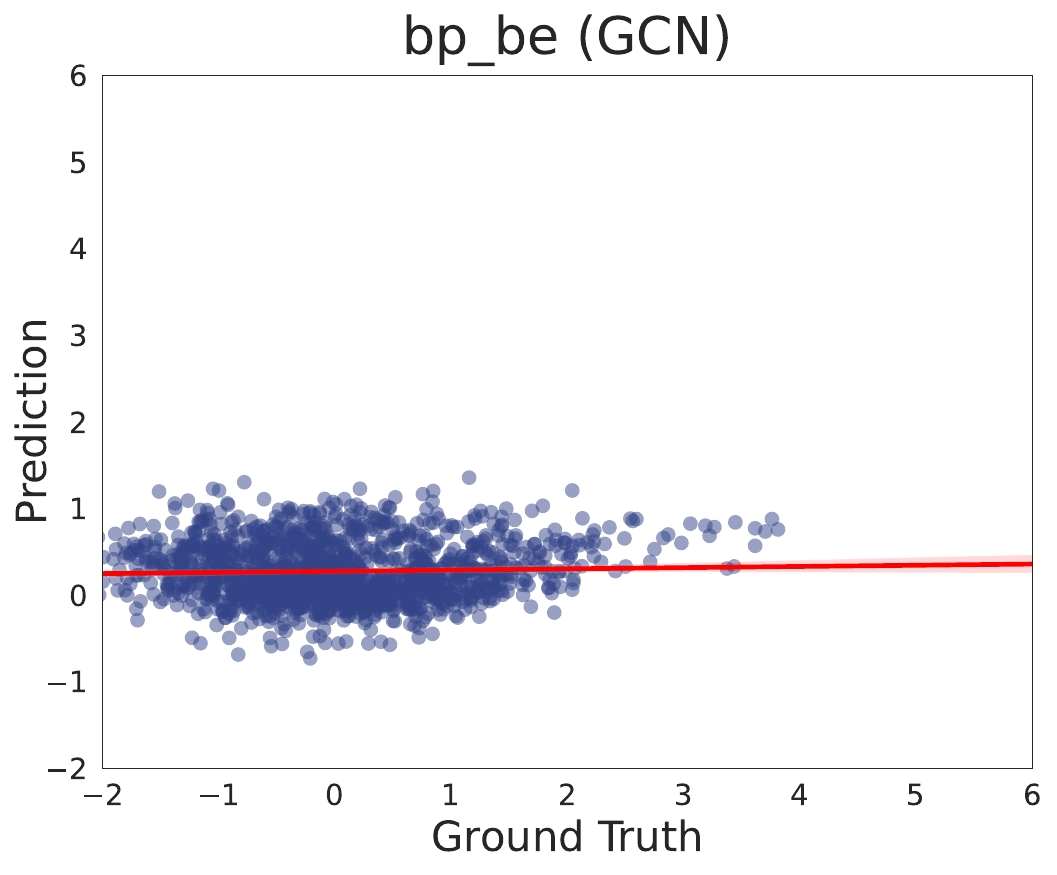}}
    \subfigure[]{%
        \includegraphics[width=0.24\textwidth]{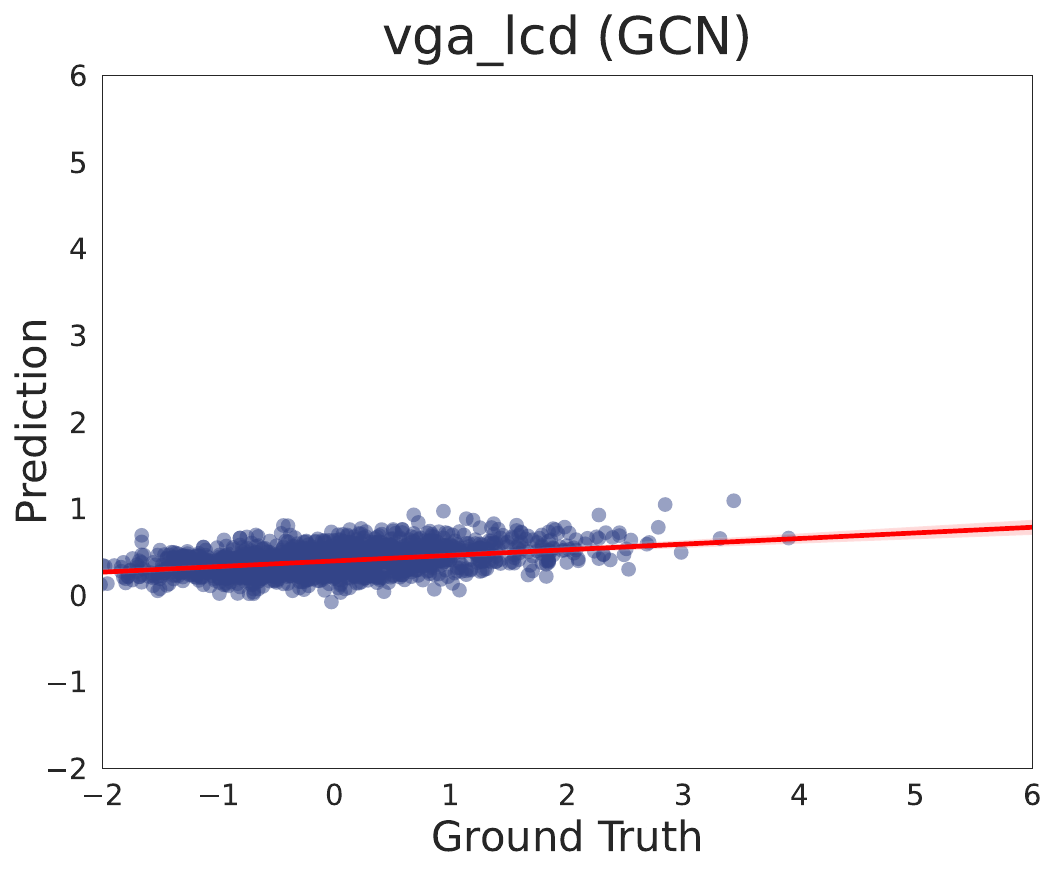}}
    \subfigure[]{%
        \includegraphics[width=0.24\textwidth]{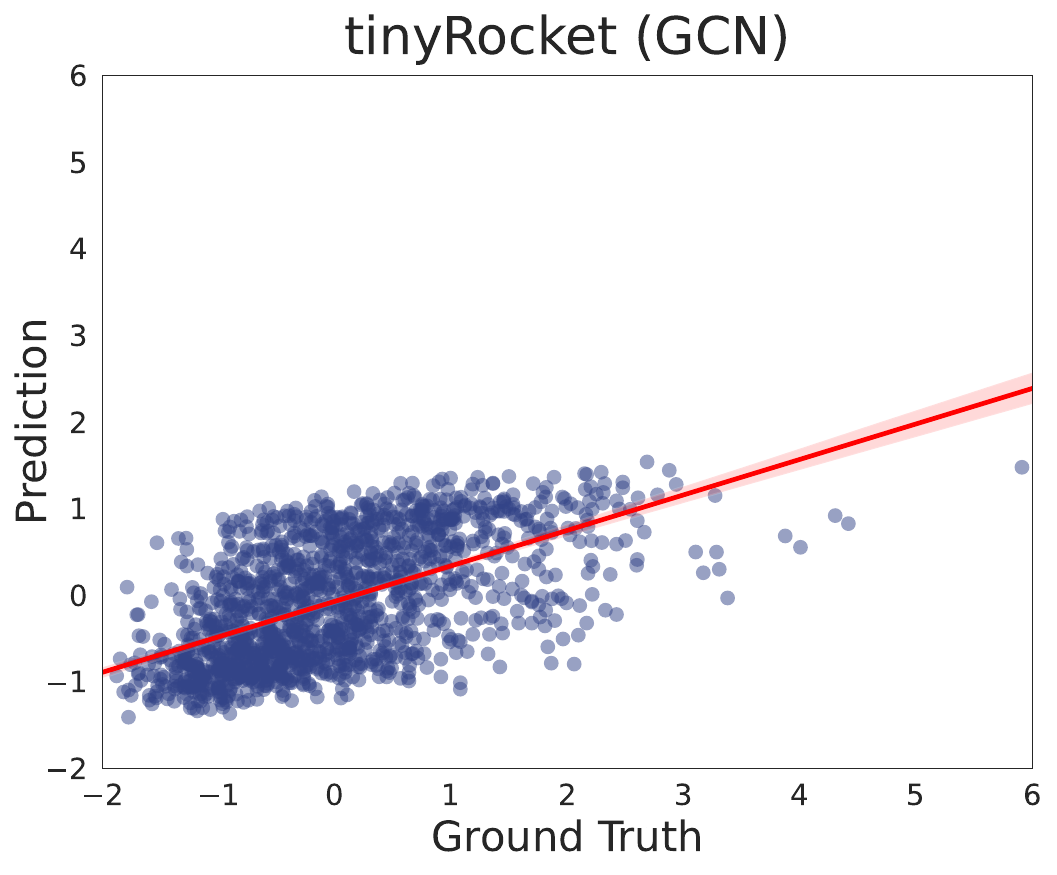}}
    \medskip
    \subfigure[]{%
        \includegraphics[width=0.24\textwidth]{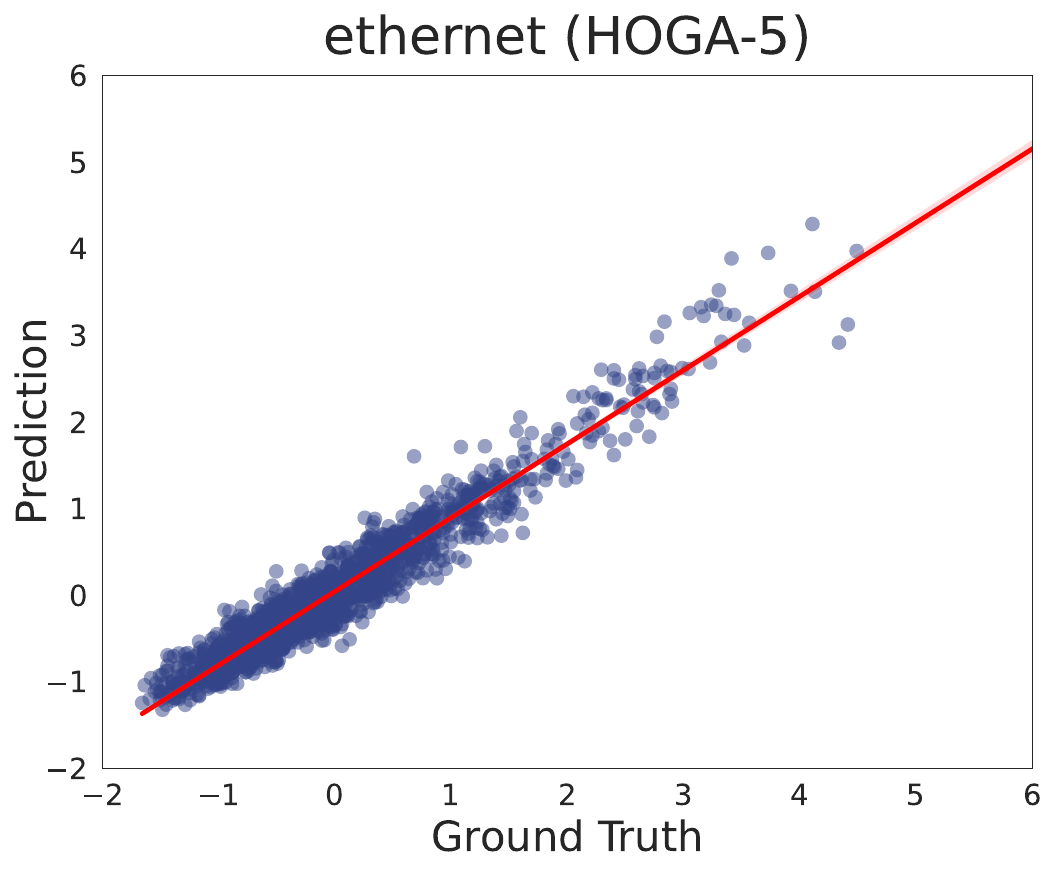}}
    \subfigure[]{%
        \includegraphics[width=0.24\textwidth]{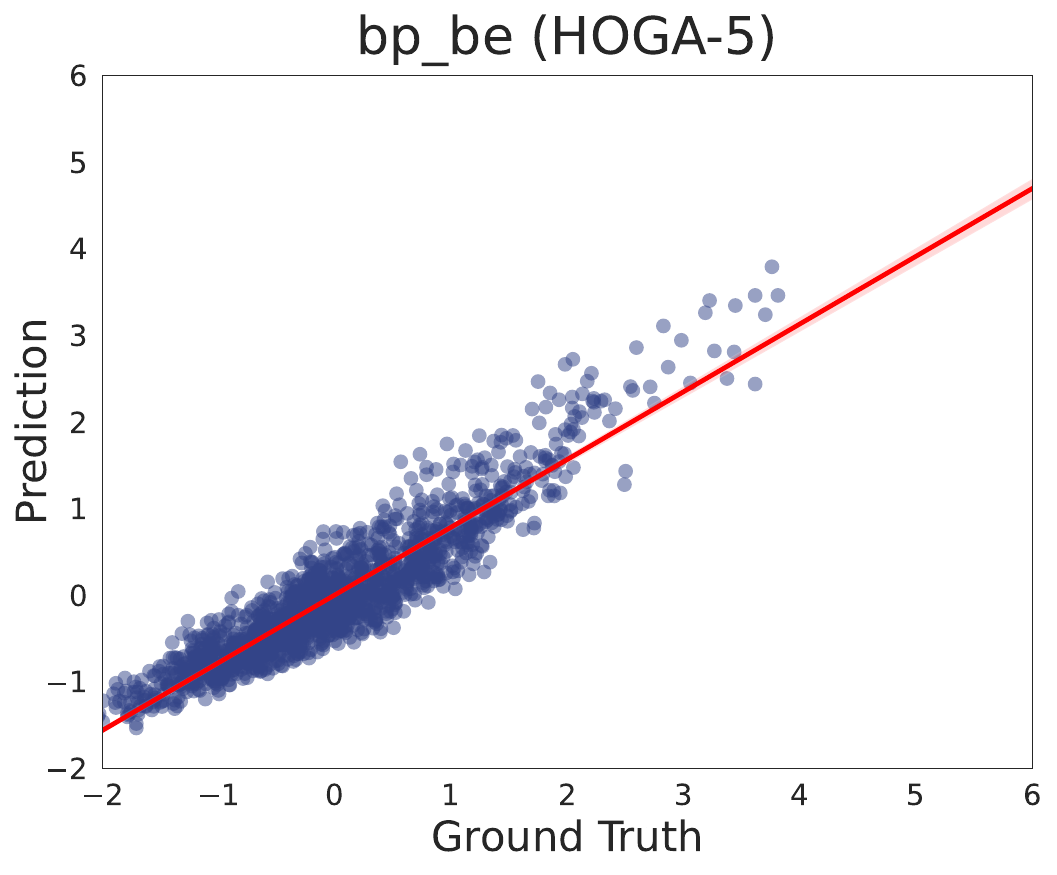}}
    \subfigure[]{%
        \includegraphics[width=0.24\textwidth]{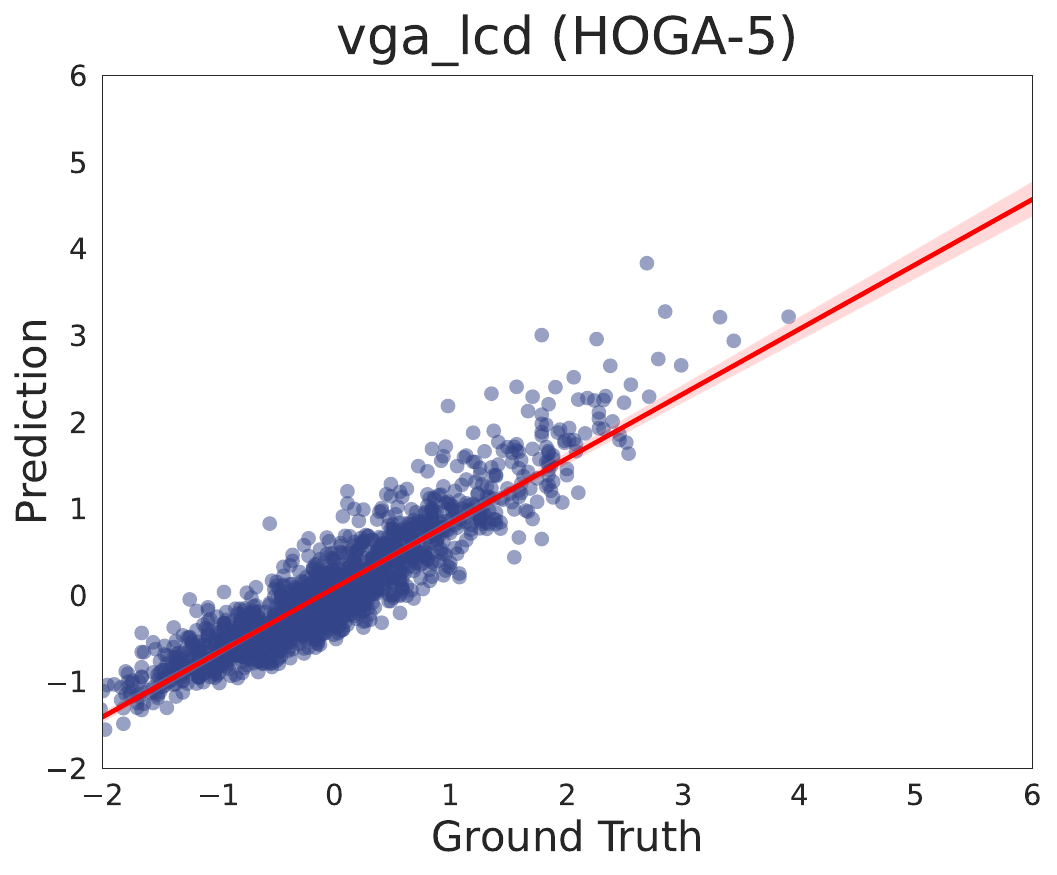}}
    \subfigure[]{%
        \includegraphics[width=0.24\textwidth]{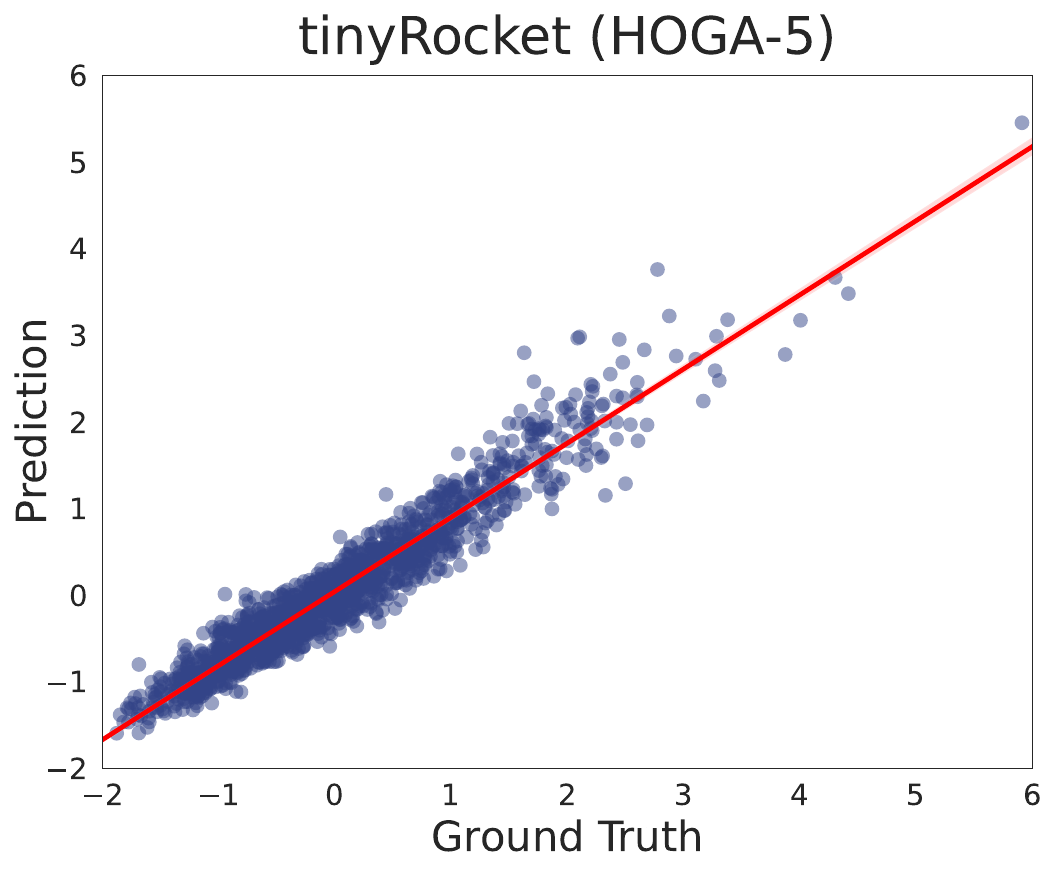}}
    \vspace{-15pt}
    \caption{GCN and \name-$5$ predictions vs. ground truth --- The ground truth value is preprocessed in the OpenABC-D benchmark.}
    \vspace{-10pt}
    \label{figure:qor}
\end{figure*}
\subsection{Evaluation on QoR Prediction}
\label{eval_qor}
Our baseline is a $5$-layer GCN, as previously used for the OpenABC-D
benchmark~\cite{chowdhury2021openabc}. 
Besides, we choose mean absolute percentage error (MAPE) as the evaluation metric, which is defined as $\text{MAPE} = \frac{1}{g}\sum_{i=1}^g |\frac{y_i-\hat{y}_i}{y_i}| \times 100\%$, where $y_i$ and $\hat{y}_i$ denote the ground truth and model prediction on the $i$-th sample (graph), respectively. Table \ref{qor} indicates that both \name models significantly outperform 
GCN across all test designs. In particular, \name-$5$ improves the estimation error over GCN on \textit{vga\_lcd} by a margin of $46.76\%$. The superiority of \name is further demonstrated in Figure \ref{figure:qor}, where the QoR predictions by \name-$5$ are highly correlated with the ground truth. In contrast, the GCN model fails to accurately predict the actual QoR values on those unseen designs. Moreover, by comparing \name-$5$ and \name-$2$, we can see the trade-off between the accuracy and training time of \name. Notably, \name-$2$ not only improves the average estimation error over GCN by $18.2\%$, but also achieves a notable $3.1\times$ training speedup on a single GPU.
\begin{figure}[ht]
\begin{center}
\vspace{-10pt}
	\includegraphics[width=0.8\columnwidth]{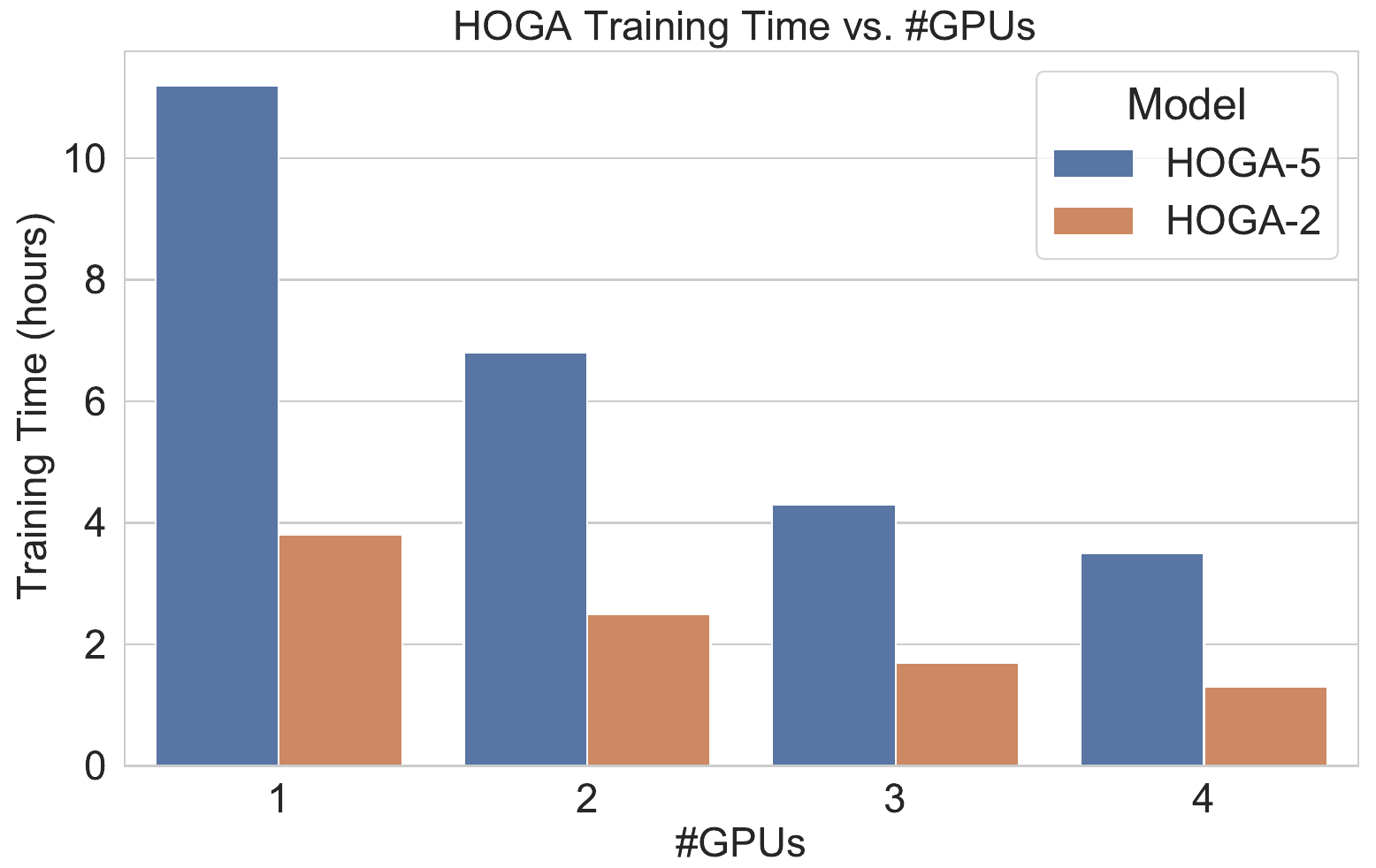}
 \vspace{-12pt}
	\caption{Multi-GPU training time of \name on OpenABC-D.} 
 \protect\label{figure:gpu}
\end{center}
\vspace{-10pt}
\end{figure}

It is noteworthy that \name can be easily accelerated through distributed training, owing to its high parallelism for learning node representations.
As shown in Figure \ref{figure:gpu}, the training time of \name almost linearly decreases when the number of GPU devices is increased. As a result, it takes only $3.5$ ($1.1$) hours to train \name-$5$ (\name-$2$) on the OpenABC-D benchmark. 
Notably, the first phase of \name for generating hop-wise features takes $13$ minutes, which is negligible compared to the overall training time. We believe that \name can be further accelerated if more computing resources are available, rendering it applicable to industrial-scale applications. 
\begin{figure}[ht]
\begin{center}
\vspace{-8pt}
	\includegraphics[width=\columnwidth]{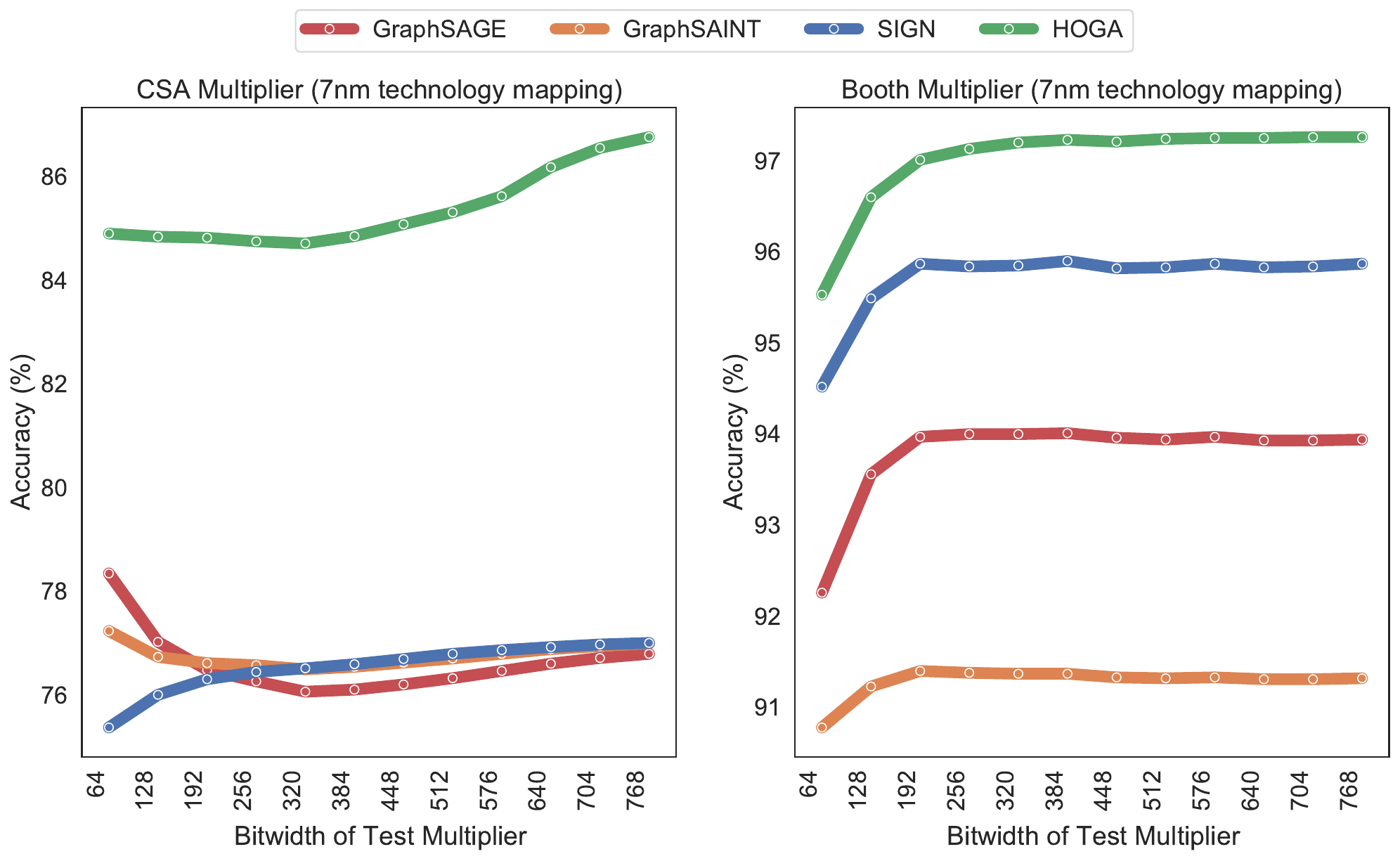}
 \vspace{-22pt}
	\caption{Functional reasoning on AIG-based CSA and Radix-$4$ Booth multipliers with $7$nm technology mapping. 
 } 
 \protect\label{figure:gamora}
\end{center}
\vspace{-18pt}
\end{figure}

\subsection{Evaluation on Functional Reasoning}
\label{eval_reason}
\begin{figure*}[ht]
\begin{center}
	\includegraphics[width=\textwidth]{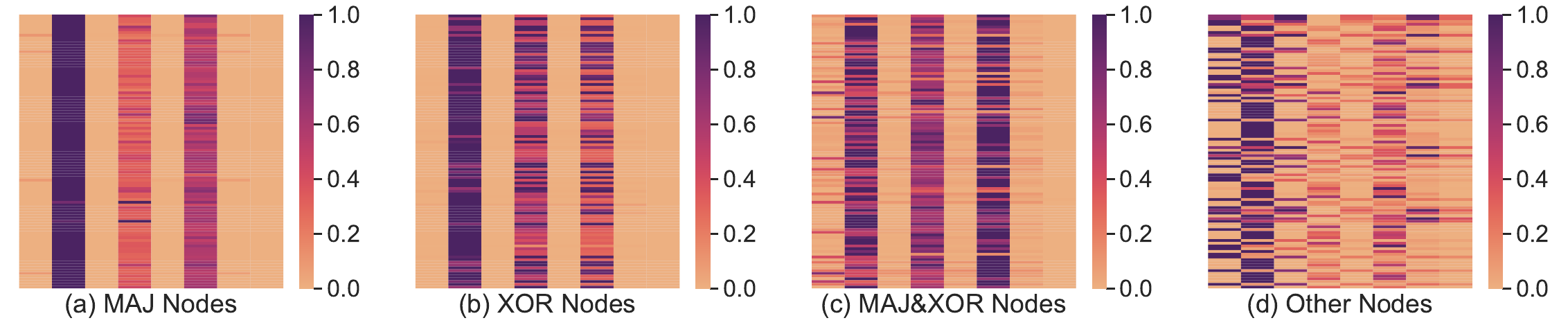}
 \vspace{-20pt}
	\caption{Visualization on hop-wise attention scores per node in the AIG-based $768$-bit Radix-$4$ Booth Multiplier --- Rows and columns in each heatmap denote nodes and their hop-wise neighbors, respectively.} 
 \protect\label{figure:visual}
\end{center}
\vspace{-10pt}
\end{figure*}
To further demonstrate the generalization capability, we evaluate \name on the functional reasoning task~\cite{wu2023gamora} by identifying root nodes of XOR and majority (MAJ) operations in AIG-based multipliers, which correspond to \textit{sum} and \textit{carry-out} of adder blocks, respectively. 
More concretely, there are $4$ node categories for classification: (1) MAJ nodes; (2) XOR nodes; (3) nodes shared by MAJ and XOR; (4) other plain nodes such as PI/PO.
Apart from choosing GraphSAGE in Gamora as our baseline, we also consider GraphSAINT~\cite{zeng2019graphsaint}, a popular sampling-based GNN model, and SIGN~\cite{frasca2020sign} that adopts an MLP model on hop-wise features.
Note that all baselines as well as \name are trained on an AIG-based $8$-bit multiplier and evaluated on multipliers with bitwidths ranging from $64$ to $768$. 

As shown in Figure \ref{figure:gamora}, all baselines perform poorly on CSA multipliers, which indicates their limited generalizability to complex designs. In addition, GraphSAINT performs even worse than GraphSAGE on Booth multipliers, indicating sampling-based GNNs are not suitable for circuit graphs, as analyzed in Section \ref{scale_motivation}. While SIGN achieves reasonable accuracy on Booth multipliers, it still largely lags behind \name on CSA multipliers. We attribute it to our proposed gated self-attention module, which adaptively learns high-order features from different hops per node. This is further confirmed in Section \ref{visual}.
As a result, \name largely surpasses baseline models on both CSA and Booth multiplier circuits, with a $10.0\%$ accuracy improvement on the $768$-bit CSA multiplier. 
More importantly, Figure \ref{figure:gamora} illustrates that the accuracy of \name exhibits a rising trend as the bitwidth of test multipliers increases, which is crucial for accurate reasoning on large-scale Boolean networks.
 

\subsection{Visualization on \name Attention Scores}
\label{visual}
For the sake of clear visualization, we randomly sample $100$ nodes per classification category from the AIG of $768$-bit Booth multiplier, based on which four heatmaps are drawn in Figure \ref{figure:visual}. Notably, the rows and columns in each heatmap correspond to nodes and their hop-wise neighbors, respectively. Each element in a heatmap row represents the hop-wise attention score $c_k$ defined in Equation \eqref{eqn:adapt_attn}, which is used to adaptively combine important hop-wise features.
Figure \ref{figure:visual} clearly shows that \name is able to identify critical features from different hop $k$ for learning the underlying Boolean functions. 
Notably, since we use a single gated self-attention layer, each $\hat{H}_k$ in Equation \eqref{eqn:adapt_attn} captures second-order interactions among hop-wise features, corresponding to second-order graph structures. As a consequence, \name skips odd-hop neighbor features and primarily focuses on $\hat{H}_k$ with $k \in \{2,4,6\}$ for learning MAJ/XOR functions, as indicated by the attention scores shown in the first three heatmaps.
As for the last heatmap, the attention scores of \name are relatively random since those nodes are PI/PO or plain AND gates, whose Boolean functions are not meaningful for this task.
\balance

\section{Conclusion}
This work introduces \name, a hop-wise attention approach for scalable and generalizable circuit representation learning. \name precomputes hop-wise features that are then fed into a gated attention module for capturing critical circuit structures. Our results showcase that \name not only comfortably scales to large-scale circuits via distributed training, but also outperforms conventional GNNs for generalizing to unseen and complex circuit designs.

\begin{acks}
This work is supported in part by a Qualcomm Innovation Fellowship, NSF Awards \#2047176, \#2118709 and \#2212371,  and ACE, one of the seven centers in JUMP 2.0, a Semiconductor Research Corporation (SRC) program sponsored by DARPA.
\end{acks}

\bibliographystyle{plain}
\bibliography{ref}


\end{document}